%% file: dshealth.tex
\documentclass[sigconf]{acmart}
\AtBeginDocument{%
  \providecommand\BibTeX{{%
    \normalfont B\kern-0.5em{\scshape i\kern-0.25em b}\kern-0.8em\TeX}}}

\acmConference[DSHealth '22]{Workshop on Applied Data Science for Healthcare
}{August 14,
  2022}{Washington, D.C.}
\usepackage{graphicx}
\usepackage{url}
\usepackage{booktabs}
\usepackage{tikz}
\usepackage{amsmath}
\usepackage{multirow}
\usepackage{subfig}
\usepackage{pgfplots}
\usepgfplotslibrary{groupplots}
\pgfplotsset{compat=1.11}
\begin{document}

\title{Towards Neural Numeric-To-Text Generation From Temporal Personal
Health Data}

\author{Jonathan Harris}
\email{harrij15@rpi.edu}
\orcid{0000-0002-8823-6602}
\affiliation{%
  \institution{Rensselaer Polytechnic Institute}
  \streetaddress{110 8th Street}
  \city{Troy}
  \state{New York}
  \country{USA}
  \postcode{12180}
}

\author{Mohammed J. Zaki}
\email{zaki@cs.rpi.edu}
\orcid{}
\affiliation{%
  \institution{Rensselaer Polytechnic Institute}
  \streetaddress{110 8th Street}
  \city{Troy}
  \state{New York}
  \country{USA}
  \postcode{12180}
}

\renewcommand{\shortauthors}{Harris, et al.}

\begin{abstract}
With an increased interest in the production of personal health technologies designed to track user data (e.g., nutrient intake, step counts), there is now more opportunity than ever to surface meaningful behavioral insights to everyday users in the form of natural language. This knowledge can increase their behavioral awareness and allow them to take action to meet their health goals. It can also bridge the gap between the vast collection of personal health data and the summary generation required to describe an individual's behavioral tendencies. 
Previous work has focused on rule-based time-series data summarization methods designed to generate natural language summaries of interesting patterns found within temporal personal health data. We examine recurrent, convolutional, and Transformer-based encoder-decoder models to automatically generate natural language summaries from numeric temporal personal health data.  We showcase the effectiveness of our models on real user health data logged in MyFitnessPal~\cite{mfp} and show that we can automatically generate high-quality natural language summaries. Our work serves as a first step towards the ambitious goal of 
automatically generating novel and meaningful temporal summaries from personal health data.
\end{abstract}




\maketitle

\section{Introduction}
\label{sec:intro}

It is now easier than ever to collect personal health data due to the increase in the production of smart devices designed to track data from multiple inputs. 
% The t
Target demographics of these products can be designated as quantified-selfers (those who maintain their own health records as a hobby), people with chronic health conditions, and everyday individuals who wish to maintain a healthy lifestyle. Quantified-selfers strive to record as much of their lives as possible using the health technologies available to them and are eager to track and learn from their own data. On the other hand, those with chronic health conditions (e.g., Type II diabetes) mainly use this information to make decisions related to future food consumption, physical activity, and so on~\cite{sundecision}. Everyday individuals who are health-conscious may also utilize a health app or device to track their progress and learn what works for them.
Unfortunately, many users of these personal health technologies tend to abandon them after a short period of time due to a lack of support when it comes to decision-making and a lack of sufficient interpretation of their data~\cite{choequantifiedself}. 
Users will then lose interest in learning from their own data and begin to record it less often. 
This results in a sparse dataset that becomes more difficult to interpret and the users end up becoming even more disengaged~\cite{codella2018}. 
Non-expert users may also incorrectly interpret their data, leading them to make unfavorable health decisions~\cite{peeldiabetes}. 
With increasingly more data collected over longer periods of time, it becomes more and more difficult to understand it. 
In light of this, there is a need for an automated system that can interpret and surface meaningful insights to aid users in their progress towards their health goals.
% Today, the most common methods utilized by everyday users to interpret their data include a search query online, conversation with a trusted health expert, or mobile apps that provide general information about their health. 
% We argue that none of these methods are sufficient to aid users on a regular basis. The search query will not be personalized to the user and will lack the context required to provide useful results. Health experts are a great resource, but a meeting with an expert depends on both the expert's availability and the user's financial stability. Current mobile apps generate summaries of a user's data; however, these summaries tend to be general summarizations that can be easily found within the data. 
This problem was partially addressed previously by works~\cite{businessdata,processes,trends,trendsapproach,kacprzyk2008linguistic,eldercare} (inspired by~\cite{prototype,yagerapproach})  designed to generate natural language summaries of temporal data using summary templates, or ``protoforms.'' 
% These approaches were inspired by~\cite{prototype,yagerapproach}. 
A protoform is essentially a summary with special ``blanks'' to be filled with specific types of words, such as summarizers (conclusive phrases), 
quantifiers (phrases that specify how often a conclusion is true), attributes (variables of interest), time windows (e.g., weeks, months), and days of the week (e.g., Friday).
The structure of an example protoform is: \textit{On $\langle$quantifier$\rangle$ $\langle$sub-time window$\rangle$ in the past $\langle$time window$\rangle$, your $\langle$attribute$\rangle$ was $\langle$summarizer$\rangle$}. This could generate the following example summary: ``On \textit{most of the days} in the past \textit{week}, your \textit{calorie intake} was \textit{high}.'' We call this a standard evaluation summary at the daily granularity.
In recent work \cite{harris2021framework}, we created a comprehensive hierarchy of twelve different protoforms to summarize different types of patterns of interest in time-series data. The summaries range from simple (e.g., standard evaluation and comparison) -- those that focus on observations that are more apparent to the everyday individual -- to more complex (e.g., if-then and cluster-based patterns) -- those that describe longer patterns discovered using more advanced data-mining techniques.
%(e.g., the SPADE frequent item-set mining algorithm~\cite{spade} and the Squeezer clustering algorithm~\cite{squeezer}). 
We use the hierarchy to generate summaries (via our summarization framework) describing behavioral patterns in real user health data. 

Although rule-based approaches can be effective, the reliance on the use of protoforms limits the diversity of the summary output. Furthermore, extending the summarization framework requires manually defining new temporal patterns (and subsequently creating new protoforms) to generate new summaries.
% they still rely on the use of protoforms which limits the diversity of the summary output. 
% At the same time, any pre-determined selection of patterns also limits the potential discovery of new patterns that may be present in the data. There is also a need to continuously develop novel data mining algorithms to find specific patterns defined by the creator. 
% The main objective of our work is to surface meaningful behavioral patterns to a non-expert user in a way that allows them to take action given the knowledge presented to them. 
In contrast, we aim to train deep learning models to both learn and fill the protoform templates presented in our framework. We believe that a transition to deep learning gives our framework more freedom to grow on a summarization and pattern mining level. Deep learning models may discover temporal patterns that we cannot see and present those patterns in natural language.
% This work is a proof of concept.
% In other words, this work can be seen as a proof of concept until a corpora is created.
% In light of this, we decide to use the summary output of \cite{harris2021framework} as our ground truth.
% Our aim is to use deep learning to bypass the need to continuously build a summarization framework and to allow the trained model to discover new temporal patterns on its own. 
% Achieving this would help expand the potential of data-to-text generation within the personal health domain.
We present an end-to-end neural approach for time-series summarization, exploring the spectrum of recurrent, convolutional, and Transformer-based models to automatically generate natural language summaries from numeric temporal personal health data. To our knowledge, this is the first such approach in the personalized health domain.
Given the lack of publicly available ground-truth summaries from personal health data, we rely on the summaries generated from our protoform-based summarization framework to train the models. We showcase summaries generated from real user data from MyFitnessPal~\cite{mfp}, and show that the automatically generated summaries are both personalized and of high quality. Our models achieve good accuracies and high BLEU scores~\cite{bleu} for many summary types. In other words, our models can effectively learn to generate understandable natural language summaries automatically from numeric time-series data. Our work should thus be considered as a proof-of-concept that opens up the tantalizing possibility of generating new temporal summary types and bypassing the need to manually extend rule-based approaches. 

%summarization framework. It should be noted %that we only train our models on the %outputs of the summarization framework %since there are no existing corpora %containing temporal health summaries that %can be used for training purposes (to the %best of our knowledge). 

% The generated summaries are both personalized and of high quality. 
% Our models achieve competitive accuracies and high BLEU scores for many summary types~\cite{bleu}.
\vspace{-0.15in}
\section{Related Work}
\label{sec:related}

According to~\citet{vanderlee}, there are three families of data-to-text generation methods: statistical machine translation~\cite{koehn,lopez,vandeemter,sanby}, neural machine translation~\cite{klein,content,ferreira,zhao,puduppully,uehara,ehr,med2vec,li}, and rule-based linguistic summarization~\cite{boran2016overview,harris2021framework,reiter}. Neural and statistical methods generally involve training models to automatically generate natural language summaries of data, while rule-based methods depend on the use of protoforms to model their summary output. There are definite benefits and drawbacks between each family, especially between the machine translation methods and the rule-based methods. 
Rule-based methods tend to have better performance and higher textual quality; however, these methods require manual creation or extension which can be considerably time-intensive. 
Most rule-based approaches find simple conclusions based on the trend/concavity of a time series and relay this to the user in a templatized natural language summary. 
In our previous work~\cite{harris2021framework}, we employed various data mining algorithms to discover hidden patterns within temporal personal health data and generated summaries via different rule-based protoforms. 
% These methods also seem to be more appropriate for simpler cases. 
They are evaluated by humans, and make use of objective measures~\cite{boran2016overview}, such as summary length and relevance. 

In the field of neural machine translation~\cite{uehara,puduppully,zhao,ferreira,content,gao,wiseman}, neural and statistical methods bypass the need for manual rule creation, but they rely on large datasets and are generally lacking in performance and text quality. 
The models' reliance on large datasets can be especially difficult in certain domains, such as in personal health. 
For evaluation, these models are typically compared using the BLEU score, which is designed to measure the agreement between the model output and the reference sentences. 
Notable examples include \citet{murakami2017} and \citet{aoki2018} who present the Market Reporter model, which can handle inter-market relationships for stock market data (e.g., relationships between stock trends for the Nikkei and Dow Jones indices). 
The authors paired time series sequences gathered from Thomson Reuters DataScope Select with associated market comments from Nikkei Quick News (NQN). The summaries generated by this model were limited to simpler conclusions, such as a continual rising trend that could be easily viewed in the data. 
In contrast to the works mentioned above, our aim is to construct neural sequence-to-sequence (i.e., numeric-to-text) generation models for temporal personal health data to generate summaries of meaningful and interesting patterns.
% % In order t
% To preserve the textual quality of the summary output of our neural models, we train them on raw temporal data paired with the summaries generated by the time-series summarization framework presented in~\cite{harris2021framework}. 
% These summaries were evaluated by a human user study and deemed to be of high quality. Thus, our models are trained to produce textual output similar to those summaries and we evaluate them via the BLEU score.

\input{learning_task}

\section{Learning Task}
\label{sec:learningtask}

Before delving into the encoder-decoder architectures, we define the learning task for numeric-to-text neural models. 
% One of the main challenges in numeric-to-text summarization in a temporal setting is the encoding task. 
% The most common sequence encoding is textual input where words in a sentence may be connected to others in a semantic manner and can be grouped by grammatical functions (e.g., noun phrases). 
% Once these semantic connections are made, we often translate the textual input to produce the same meaning in another vocabulary (e.g., English-to-German translation). 
% For time-series data, we deal with sequences in different temporal granularities (e.g., weeks) where one day may not be related to another unless we give it meaning, such as a relative comparison or a pattern within a time window. 
% Once these concepts are established, we encode them in a way to generate natural language text describing said concepts. 
% In order to capture the temporal patterns within time-series data and generate relevant natural language, there is a need to modify standard sequence-to-sequence model architectures to effectively encode time-series data.
A main challenge is the lack of suitable ground-truth training data pairing personal health data with high-quality summaries that can be used for training. On the other hand, we do have relatively high-quality summaries from our recently proposed summary type hierarchy. We also conducted a user study to evaluate the output summaries
% these personalized summaries have been evaluated by humans via a user study.
by their readability, their comprehensiveness, their usefulness, and how well they align with the data they are describing.
% Each summary type contains a summarizer (e.g., `high', `low') that describes a time window in the user's data relative to the user's personal history. In other words, the summarizer is personalized to the user: `high' for one user could be `low' for another. Most of the summary types also contain quantifiers which specify how often the summarizer is true within the time window. 
% The summaries deemed most useful were the standard/goal evaluation summaries, the comparison summaries, and the day-based pattern summaries. 
Thus, given the lack of publicly available domain expert summaries for personal health data, as a first step, we use the summaries produced from our rule-based framework as the ground truth to train our neural models. We believe this is an effective strategy since we can train our models on a variety of summary types, establishing a suitable state-of-the-art method for this task. Further, this also showcases the proof of concept, that it is indeed possible to automatically generate high-quality natural language summaries from numeric data using deep learning models. In the future, our aim is to generate free-form summaries.
% In the future, we plan to generalize our models to generate multiple summary types, as well as generate new types of summaries.

The learning task is to translate raw or numeric time-series subsequences into natural language summaries, as reflected in Figure \ref{fig:learning_task}. Here, the input is numeric time-series data comprising the subsequences comprising the past week (top) and the entire user history (bottom). The neural network models are then expected to generate a natural language summary, as shown.
% We focus our model on a single univariate summary type at a time for an individual at a personalized level. 
% Table \ref{tab:summaries} contains the summary types that fit our learning task along with an example protoform and a resulting summary output. 
Our models receive training pairs containing a time series subsequence of personal health data (e.g., calorie intake), the natural language summary generated for it, and the associated protoform for that summary. The summary type is selected prior to training and the learning models are evaluated based on their accuracy and BLEU score for each summary type.

\vspace{-0.075in}
% \section{Numeric-to-Text Models}
\section{Numeric-to-Text Models}
\label{sec:arch}

% ,
We introduce CNN-LSTM, Transformer, and Transformer-LSTM encoder-decoder models for numeric-to-text translation. 
%\paragraph{Model Input:}
The input to all three models comprises the 
the short-term ($x_{short}$) and long-term ($x_{long}$) representations of the temporal personal health data. In our case, the short-term representation of the data is the input time series subsequence of interest (shown on top left in Fig.~\ref{fig:learning_task}), while the long-term representation is the entire time series (shown on bottom left). 
% $x_{short}$ is the subsequence for which a summary is sought, whereas $x_{long}$ provides the context for each user. 
Formally, we define the long-term representation as $x_{long} = (x_1, x_2, ..., x_N)$ where $x_i \in \mathbb{R}$ and $N$ is the length of the entire time series, and the short-term representation as $x_{short} = (x_i, x_{i+1}, ..., x_j)$ where $1 \leq i,j \leq N$ and $i < j$. The length of $x_{short}$ depends on the summary type the model is learning.
Since we are working with personalized summaries (e.g., medium sodium intake for one user can be high intake for another user), they require the context of the time series ($x_{long}$) to be useful. 
% We also require $x_{short}$ to provide information of the granularity of the time window ($TW$) we are working with, as well as the part of the time series the model should be looking at. 

% \subsection{CNN-LSTM Model}
% The architecture for the CNN-LSTM model can be found in Figure \ref{fig:cnnlstm}. 

% The types of patterns we are interested in are not meaningful to every user but they are useful to the current user. 
% Also, as mentioned in Section \ref{sec:learningtask}, the summaries to be learned contain words such as `high' or `low', which are based on the user's personal history. The model requires the input of the entire time series to fully understand what `high' and `low' refer to in that context. 
% The CNN-LSTM's encoder is CNN-based.
\input{output}
For the CNN-LSTM model, we feed the two representations of the input data into separate, yet similar, convolutional encoder layers and concatenate the resulting hidden states with the original $x_{short}$ and $x_{long}$ sequences before sending them through fully-connected dense and dropout layers. 
%The layer outputs are then sent through a series of fully-connected layers for the decoding step. 
For the decoding step, we utilize two separate LSTM decoders: a summary decoder and an additional template decoder. The summary decoder generates the predicted summary tokens $y_{pred} = ``s_{1}\;s_{2}\;...\;s_{n}"$
where $n$ is the number of tokens generated by the LSTM for the resulting natural language summary, while the template decoder generates the predicted template tokens $y_{proto} = ``t_{1}\;t_{2}\;...\;t_{n}"$ for the resulting protoform. These template tokens are generated directly from the summary tokens $y_{pred}$ for input. It may seem that the same $y_{proto}$ will be fed as input for each example; however, any summary type capable of generating summaries that vary in length (e.g., if-then pattern summaries) will have varying inputs for $y_{proto}$. Summary tokens $y_{pred}$ and template tokens $y_{proto}$ are the two outputs of our model. In essence, the model has two similar learning tasks: the translation of a time series sequence with added context to a natural language summary and its associated protoform.
% , as shown in Fig.~\ref{fig:output}.
% We added the template decoder with the knowledge of the structured input summaries paired with the time series sequences. 
%The CNN-LSTM model is trained on a single summary type at a time so each input summary has a similar structure. 
Whereas we are mainly interested in the summary output, the template decoder allows the model to learn the protoform structure which results in better summary output. Once it learns the protoform using the input template tokens, it can automatically determine what the ``blanks'' should be. 
%We help the model learn the template by feeding it a templated version of the summary tokens, which represents the protoform structure. Each special ``blank'' in the protoform is represented with a special token representing the type of word that should fill in the blank. 
For example, given the set of template tokens ``In the past full \textbf{TW}, your \textbf{A A} has been \textbf{S},'' it can generate a summary such as ``In the past full \textbf{week}, your \textbf{calorie intake} has been \textbf{moderate}.''
The template tokens help the neural network focus on the special ``blanks'' mentioned in Sec.~\ref{sec:intro}, whereas the summary tokens can focus on the final token-level natural language summary. 
% These special tokens are listed in Table \ref{tab:specialtokens}. 
% Every other token remains untouched.
% \vspace{-0.025in}
% \begin{table}[!ht]
%     \centering
%     \small
%     \begin{tabular}{|c|c|}
%         \hline
%         Token Type & Token \\\hline
%         Summarizer & S \\\hline
%         Quantifier & Q \\\hline
%         Time Window & TW \\\hline
%         Day of the Week & D \\\hline
%     \end{tabular}
%      %\vspace{0.1in}
%     \caption{Special Tokens in Template Token Input}
%     \label{tab:specialtokens}
%     \vspace{-0.305in}
% \end{table}
% The summary decoder uses a negative log-likelihood function based on the summary decoder output and the expected summary tokens.
% We also feed the template decoder output into a negative log-likelihood loss function and multiply the resulting value by the number of tokens within the special blanks that are incorrect. The final value is then added to the loss from the summary decoder. 
The decoding process is shown in Figure~\ref{fig:output}. 
% The LSTM cells in both decoders generate one token at a time using the previous cell's decoding output and the previously generated token. The process is started with the final hidden state of the encoder and a start-of-sentence token ($<$s$>$).
The model utilizes a cross-entropy loss with respect to the ground-truth summary and template tokens at each position, which yields the combined loss for the summary and template decoder output. The resulting loss function given as: $L(\hat{y}_s,\hat{y}_t) = \sum_{i=1}^{n} CE(y_{s_i},\hat{y}_{s_i}) + m\sum_{i=1}^{n}CE(y_{t_i},\hat{y}_{t_i})$,
where $CE$ is the cross entropy loss per token, $n$ is the summary length, $y_{s_i}$ and $\hat{y}_{s_i}$ represent the actual and predicted summary tokens from the summary decoder, $y_{t_i}$ and $\hat{y}_{t_i}$ represent the actual and predicted template tokens from the template decoder, and $m$ represents the number of incorrect ``blanks'' in the template decoder output (i.e., $m$ provides a higher penalty).
%for token position $i$. 
% $m_i$ serves as a multiplier to add extra penalty to an incorrect sentence structure.

% \input{transformer}

% \begin{figure}[!h]
%   \centering
%   \includegraphics[width=\linewidth]{transformer.png}
%   \caption{Architecture of the Transformer model. The general architecture is inspired by~\cite{attention} and an implementation of a Transformer encoder optimized specifically for handling time series (found here: \url{https://github.com/maxjcohen/transformer}). Each ``N$\times$'' or ``$\times$N'' indicates a stack. The leftmost stack is the time-series Transformer (TST) stack, while the remaining stacks are the summary and template decoder stacks from left to right. The TST encoder stack processes the raw input data with a positional encoding and its outputs are sent to the Multi-Head Attention layers of the decoder stacks, as well as the output embeddings. From there, the decoder stacks use the encodings to generate output probabilities of the summary and template tokens. }
%   \label{fig:transformer}
% \end{figure}
% \vspace{-0.05in}
% \subsection{Transformer Models}
% \subsubsection{Time Series Transformer-Transformer Model}
% The architecture for the Transformer model is shown in Figure \ref{fig:transformer}. 
Transformers are a viable alternative to recurrent and convolutional networks via their use of attention; therefore, we decided to test the summary generation task on a numeric-to-text Time Series Transformer-Transformer model. The original Transformer~\cite{attention} focuses on text-to-text machine translation. Thus, we replace the text encoder with one that can process numeric time-series data. We extend the Time Series Transformer (TST)~\cite{cohen} encoder, and pair it with a Transformer decoder (for natural language generation) to construct a model for numeric-to-text generation. The input to TST encoder is the concatenation of $x_{short}$ and $x_{long}$, and it 
utilizes multi-head attention by dividing the queries, keys, and values into chunks using a moving window (we use window size 12).
%and we do not use teacher forcing. 
For decoding, we employ dual Transformer decoders to train the model on both the protoform structure and natural language so that it can produce a more comprehensive output. Teacher forcing is not used during training. 
We also experimented with the TST encoder and an LSTM decoder model. 
% The architecture for the Transformer-LSTM model is shown in Figure \ref{fig:transformer_lstm}. 
We hypothesized that the LSTM decoder could be a possible alternative to the Transformer decoder, especially when receiving encodings from time-series data since the Transformer decoder may not be the ideal pairing for the TST encoder. 
% Although this model does not take advantage of the masked multi-head attention of the Transformer decoder, the LSTM still utilizes the self-attention from TST for its own attention vector within the architecture. 
The encoder-decoder connection between the TST and LSTM is similar to that of the CNN-LSTM model. 
\vspace{-0.05in}
\section{Experiments}
\begin{table*}[!t]
    \centering
    % \vspace{-0.1in}
    \footnotesize
    \begin{tabular}{|c|c|c|c||c|c|c|}
        \hline
        \multirow{2}*{Summary Type}  & \multicolumn{3}{|c||}{Accuracy} & \multicolumn{3}{|c|}{BLEU Score}\\\cline{2-7}
        & CNN-LSTM & TST-Transformer & TST-LSTM & CNN-LSTM & TST-Transformer & TST-LSTM\\\hline
        Standard Evaluation (TW granularity) & \textbf{1} & 0.98 & \textbf{1} & 0.9999 & 0.998 & \textbf{1}\\\hline
    Standard Evaluation (sTW granularity) & \textbf{1} & 0.96 & \textbf{1} & 0.999 & 0.996 & \textbf{0.9998}\\\hline
    Day-Based Pattern & \textbf{1} & 0.846 & \textbf{1} & 0.9998 & 0.987 & \textbf{0.9999}\\\hline
    Goal Evaluation & \textbf{0.98} & 0.5 & 0.92 & \textbf{0.997} & 0.954 & 0.991\\\hline
    Goal Assistance & 0.86 & 0.745 & \textbf{0.87} & 0.854 & 0.778 & \textbf{0.866}\\\hline
    Standard Trend & \textbf{1} & 0.29 & \textbf{1} & \textbf{0.9999} & 0.919 & \textbf{0.9999}\\\hline
    If-Then Pattern & \textbf{1} & 0.998 & \textbf{1} & \textbf{0.9999} & \textbf{0.9999} & 0.9998\\\hline
    Day If-Then Pattern & 0.1 & 0.07 & \textbf{0.14} & 0.845 & \textbf{0.955} & 0.853\\\hline
    Evaluation Comparison & \textbf{0.97} & 0.8 & \textbf{0.97} & \textbf{0.99} & 0.968 & \textbf{0.99}\\\hline
    Goal Comparison & \textbf{0.97} & 0.59 & 0.73 & \textbf{0.994} & 0.953 & 0.944\\\hline
    Cluster-Based Description & 0.43 & 0.74 & \textbf{0.97} & 0.894 & 0.98 & \textbf{0.995}\\\hline
    Cluster-Based Pattern & 0.43 & 0.26 & \textbf{0.71} & 0.861 & 0.925 & \textbf{0.931}\\\hline
    Standard Pattern & \textbf{0.85} & 0.3 & 0.82 &\textbf{ 0.977} & 0.915 & 0.961\\\hline\hline
   \textbf{Average} & 0.815 & 0.621 & \textbf{0.856} & 0.955 & 0.948 & \textbf{0.964}\\\hline
    \end{tabular}
    % \vspace{0.1in}
    \caption{Experiment Results: Comparing the Numeric-to-Text Encoder-Decoder Models}
    \label{tab:modelresults}
    \vspace{-0.3in}
\end{table*}
The models were trained using PyTorch, on a Linux-based machine with an NVIDIA Tesla V100 GPU. For reproducibility purposes, our open source implementation is available from~\url{https://github.com/neato47/Neural-Numeric-To-Text-Generation}.
We conducted our experiments using the MyFitnessPal food log dataset~\cite{mfp},
%\footnote{Scraped by researchers from the Singaporean Management University from a website containing publicly accessible MyFitnessPal food diary pages.}
which contains 587,187 days of real food log data across 9.9K users (389 of them were selected), each tracking up to 180 days worth of food and nutrient intake data.
Users were expected to log the food items they consumed and their daily calorie goals, while the MyFitnessPal database added in the associated nutrient information and total daily intake. 
We train our models on each summary type separately and evaluate their performance using the BLEU score and the model's prediction accuracy. The accuracy is determined by how exactly each summary in the predicted output matches the expected output on a token-to-token basis. 
In terms of hyperparameters, we used the Adam optimizer with a learning rate of 0.0001 and cross-entropy loss for all three models.
%as well as negative log-likelihood loss for the loss function. 
For the CNN-LSTM model, the hidden encoder/decoder size is 180 and the encoder's output size is 256. 
%Our CNN uses the hyperbolic tangent function as its activation function. 
The CNN kernel size is $1 \times 3$, with a stride of 1 and padding of 1 for both convolutional layers.
%(one set of convolutional layers for the input sequence and series as shown in Figure~\ref{fig:cnnlstm}). 
The max pooling layers
%in the CNN-LSTM architecture 
have a kernel size of 2 and a stride of 2. Only one linear layer is used before the output neurons. The output dimension of the decoder is the length of the largest ground-truth summary. The CNN-LSTM model is trained in batches of size 180 for 78 epochs. % (with some exceptions, which are explained in Section \ref{sec:training}).
For the Transformer-based models, the input embeddings are 64 dimensional ($d_{model}$), with query, key and value dimensionality of 8, with 4 heads. There are four stacks encoder and (summary and template) decoder layers. A dropout probability of 0.2 is used for both the encoder and decoder layers. The TST-LSTM model was trained in batches of size 8 for 30 epochs.

% As can be seen, the Transformer-LSTM model proves to a better fit for the personal health domain.
% \begin{table}
%     \centering
%     \small
%     \begin{tabular}{|c|c|c|}
%         \hline
%         Summary Type  & Calorie & Carbohydrate\\\hline
%         Standard Evaluation (TW granularity) & 1 & 0.96\\\hline
%   Standard Evaluation (sTW granularity)& 1 &1 \\\hline
%   Day-Based Pattern & 1 & 0 \\\hline
%     Goal Evaluation & 0.98 & 0.98 \\\hline
%     Goal Assistance & 0.86 & 0.89  \\\hline
%     Standard Trend & 1 & 1\\\hline
%   If-Then Pattern & 0.99 & 0.32 \\\hline
%     Day If-Then Pattern & 0.1 & 0.02 \\\hline
%     Evaluation Comparison & 0.97 & 0.98 \\\hline
%     Goal Comparison & 0.97 & 0.97 \\\hline
%     Cluster-Based Description & 0.61 & 0.52 \\\hline
%     Cluster-Based Pattern & 0.46 & 0.69  \\\hline
%     Standard Pattern & 0.85 & 0.9 \\\hline
%     \end{tabular}
%     %\vspace{0.1in}
%     \caption{CNN-LSTM model trained on calorie intake data tested on carbohydrate intake data}
%     \label{tab:accuracies_cnnlstm}
%     \vspace{-0.3in}
% \end{table}
%We train our models on each summary type separately and evaluate their performance using the BLEU score and the model's prediction accuracy. The prediction accuracy is determined by how exactly each summary in the predicted output matches the expected output on a token-by-token basis. 
% In other words, 
% $\text{accuracy} = \dfrac{\text{\# of matching summaries}}{\text{Total \# of summaries}}$.

% \smallskip
% \noindent
% {\bf Model Comparison:}
We ran experiments on the users' calorie intake data; the comparative results for the three models, for each summary type, are reported in Table~\ref{tab:modelresults}. 
The CNN-LSTM's average prediction accuracy across all of the summary types is around 0.814, the TST-Transformer's average accuracy is around 0.621, and the TST-LSTM's is around 0.856. The TST-LSTM model also has the highest exact match accuracy for 10 out of the 13 summary types. 
The BLEU score~\cite{bleu} measures the agreement between the model output and the reference sentences by calculating the n-gram overlap between the output and reference sentences. A score of 1 indicates identical sentences.
The CNN-LSTM model has an average BLEU score of 0.955, the TST-Transformer model has an average of 0.948, and the TST-LSTM model has an average of 0.964. The TST-LSTM model also has the highest BLEU score for 9 out of the 13 summary types. Based on average accuracy and BLEU score alone, the TST-LSTM model performs better when it comes to matching the exact summary output and it makes predictions that are closer to the target summary output more often. This shows that the TST-LSTM model is the better model.
% , and interestingly, it performs better than the TST-Transformer approach. 
% The results with 100\% accuracy can indicate that the models are overfitting on those datasets. To evaluate this possibility, we tested our trained models on carbohydrate intake data to see how they would perform on entirely new data with a different variable. We believe this is a sufficient test since a daily calorie intake is typically around 2,000 on average, while carbohydrate intake is typically around 200-300 grams. One could argue that the calorie and carbohydrate intake variables can be correlated so it would be easier to detect the patterns after data normalization; however, that depends on the diet and every user in our dataset is different. The CNN-LSTM results are found in Table \ref{tab:accuracies_cnnlstm}. For the interest of space, the Transformer-based results for carbohydrate intake were not included; however, the conclusion remains the same. Most of the summary types have no significant differences in accuracies with the new dataset; however, the if-then and day-based pattern summaries may be overfitting due to their drops in accuracy. It is interesting, however, to see increases in accuracy for some of the summary types when tested on the carbohydrate intake data. For the most part, these results show the robustness of our approach. 
Looking at the summary types, it seems that the models had the most trouble with day if-then pattern, goal comparison, cluster-based pattern, and standard pattern summaries. 
Please refer to~\cite{harris2021framework} for more information on these summary types.
All three models mainly struggled to correctly guess the days of the week (e.g., Friday) for the day if-then pattern summaries. It may be difficult to keep track of the days based on the data.
Goal comparison summaries compare a user's adherence to a goal between two time windows at the weekly granularity. It appears that the TST-Transformer had trouble factoring in the calorie intake goal for the comparison, which may point to the raw input. It only had an accuracy of 0.59 for this type, while it had an accuracy of 0.8 for evaluation comparison summaries. 
Standard trend summaries describe how often a time series changes slope from one day to the next; however, the CNN-LSTM struggles for this summary type with an accuracy of 0.29. It is possible that the CNN encoder is having trouble detecting the change in slope. 
Cluster-based pattern summaries explain what happened directly after weeks that are most similar to the most recent week $w$. This information helps predict what could happen in $w'$, the week after week $w$. The cluster-based description summary type is a description of the similar week that is most recent.
The $x_{short}$ of both summary types is the most recent week. This may hinder the CNN-LSTM's and TST-Transformer's ability to find the connections between the most recent week and the weeks similar to it since the CNN-LSTM only had an accuracy of 0.43 for both summary types, while the TST-Transformer had an accuracy of 0.26 for the cluster-based pattern summary type. It may be beneficial to add more information to the $x_{short}$ (i.e., similar weeks and the weeks after them).
The standard pattern summary type is very similar to the cluster-based pattern summary type, except it only uses the most recent similar week to predict the user's behavior in week $w'$ and its $x_{short}$ contains the most recent similar week, the week directly after, and $w$. The CNN-LSTM also struggled with this summary type, resulting in an accuracy of 0.3. 
% Thus, it is necessary to revisit this and provide more information (namely, the similar weeks ${w}_i$ and the successive $w_{i+1}$) for this summary type. In hindsight, it is insufficient to train the models with the current $x_{short}$ for the cluster-based summary types, as this does not provide enough information to the CNN-LSTM and the TST-Transformer about what weeks they should focus on when looking at the entire time series. Without the information of the similar weeks, they may have failed to find the patterns needed to perform well. 

\vspace{-0.05in}
\section{Conclusion}
In this paper, we present and compare neural numeric-to-text machine translation models designed to translate raw temporal personal health data into natural language summaries. With these models, we surface hidden, meaningful patterns in a user's personal health data and provide them with the knowledge required to work closer to their health goals. 
% , especially for the goal-based and day if-then pattern summaries. 
This work is a proof-of-concept demonstrating the feasibility of generating explanations and summaries from personal health data. %Ideally, the system would need to be tested on a corpora containing domain expert summaries describing temporal personal health data is created. 
% In the meantime, we will continue to work with summaries generated by our summarization framework.
For future work, we plan to
% try out new architectures including models such as the T5~\cite{t5}, BART~\cite{bart}, and GPT-3~\cite{gpt3} models. We will also 
% incorporate deep neural network inspection to analyze how our models find patterns in time-series data. A further goal is to 
construct joint models that can be trained on all of the summary types at once. We also plan to explore generative models~\cite{t5,bart,gpt3} to generate novel summaries from time-series data using machine translation. Finally, we wish to look more into how we could make real-life applications of our work despite limited training data.
% , we would possibly need to gather our own natural language summaries from health data experts to enhance the summary generation of our models. 

\bibliographystyle{ACM-Reference-Format}
\bibliography{dshealth}

\end{document}

%% file: learning_task.tex
\begin{figure}[!ht]
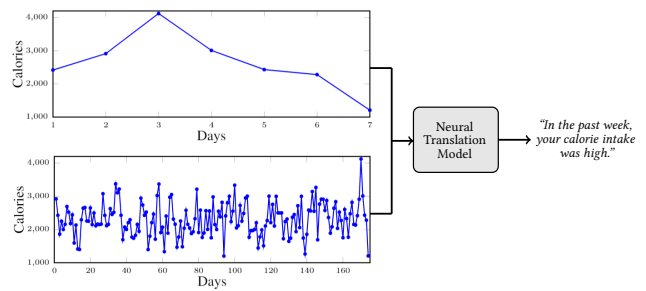

    \centering
    \vspace{-0.1in}
    \resizebox{1.0\columnwidth}{!}{
        \begin{tikzpicture}
    
        \node[inner sep=0pt] at (-1,-1)
        {\includegraphics[width=\columnwidth]{sequence.pdf}};
        \node[inner sep=0pt] at (-0.9,-4.3)
        {\includegraphics[width=1.03\columnwidth]{series.pdf}};
        
        \draw[very thick] (3.05,-0.85) -- (3.55,-0.85);
        \draw[very thick] (3.55,-0.85) -- (3.55,-4.2);
        \draw[very thick] (3.05,-4.2) -- (3.55,-4.2);
        % \draw[very thick] (4.5,-4.9) -- (4.5,-2.5);
        \draw[->,very thick] (3.55,-2.525) -- (4.05,-2.525);
        % \draw[->,very thick] (3.67,0.3) -- (5.25,-2.5);
        % \draw[->,very thick] (3.67,-4.9) -- (5.25,-2.5);
        
        \draw[fill=gray!20!white,rounded corners,thick] (4.05,-1.75) rectangle (6,-3.25);
        \node[align=center,scale=0.8] at (5,-2.5) {\Large Neural\\\Large Translation \\\Large Model};
        
        \draw[->,very thick] (6,-2.5) -- (6.75,-2.5);
        
        % \draw [fill=white,rounded corners,thick] (8,-1.75) rectangle (12,-3.25);
        \node[align=center,scale=0.8] at (8,-2.5) {\Large \textit{``In the past week,}\\\Large \textit{your calorie intake}\\\Large \textit{ was high.''}};
       
        \end{tikzpicture}
    }
    \vspace{-0.25in}
    \caption{\small Learning Task Overview: A subsequence (top left) and the entire time series of a user's calorie intake (bottom left) are fed as input into a neural translation model, which outputs a natural language summary describing a pattern or trend in the numeric personal health data. }
    \label{fig:learning_task}
    \vspace{-0.2in}
\end{figure}

%% file: output.tex
\begin{figure}[!htb]
    \centering
    \vspace{-0.2in}
    \resizebox{!}{1.4in}{
    
    \begin{tikzpicture}
    %% Decoder 1 %%
    \node at (1.2,-6.5) {\LARGE $<$s$>$};
    \draw[->] (1.2,-6.7) -- (1.2,-7);
    \draw [fill=cyan!30!white] (0.9,-8) rectangle (1.5,-7);
    \draw[->] (1.55,-7.5) -- (2.15,-7.5);
    \draw[->] (1.2,-8) -- (1.2,-8.25);
    \node at (1.2,-8.65) {\LARGE \textit{In}};
    
    \draw[thick,dotted,->] plot [smooth, tension=1.25] coordinates { (1.2,-8) (1.6,-8.1) (1.9,-6.9) (2.3,-7)};
    \draw [fill=cyan!30!white] (2,-8) rectangle (2.6,-7);
    \draw[->] (2.65,-7.5) -- (3.1,-7.5);
    \draw[->] (2.3,-8) -- (2.3,-8.25);
    \node at (2.3,-8.62) {\LARGE \textit{the}};
    
    \draw[thick,dotted,->] plot [smooth, tension=1.25] coordinates { (2.3,-8) (2.7,-8.1) (3,-6.9) (3.4,-7)};
    \draw [fill=cyan!30!white] (3.1,-8) rectangle (3.7,-7);
    \draw[->] (3.75,-7.5) -- (4.2,-7.5);
    \draw[->] (3.4,-8) -- (3.4,-8.25);
    \node at (3.4,-8.7) {\LARGE \textit{past}};
    
    \draw[thick,dotted,->] plot [smooth, tension=1.25] coordinates { (3.4,-8) (3.8,-8.1) (4.1,-6.9) (4.5,-7)};
    \draw [fill=cyan!30!white] (4.2,-8) rectangle (4.8,-7);
    \draw[->] (4.85,-7.5) -- (5.3,-7.5);
    \draw[->] (4.5,-8) -- (4.5,-8.25);
    \node at (4.5,-8.7) {\LARGE \textit{full}};
    
    \draw[thick,dotted,->] plot [smooth, tension=1.25] coordinates { (4.5,-8) (4.9,-8.1) (5.2,-6.9) (5.6,-7)};
    \draw [fill=cyan!30!white] (5.3,-8) rectangle (5.9,-7);
    \draw[->] (5.95,-7.5) -- (6.4,-7.5);
    \draw[->] (5.6,-8) -- (5.6,-8.25);
    \node at (5.6,-8.7) {\LARGE \textit{week,}};
    
    \draw[thick,dotted,->] plot [smooth, tension=1.25] coordinates { (5.6,-8) (6,-8.1) (6.3,-6.9) (6.7,-7)};
    \draw [fill=cyan!30!white] (6.4,-8) rectangle (7,-7);
    \draw[->] (7.05,-7.5) -- (7.5,-7.5);
    \draw[->] (6.7,-8) -- (6.7,-8.25);
    \node at (6.7,-8.77) {\LARGE \textit{your}};
    
    \draw[thick,dotted,->] plot [smooth, tension=1.25] coordinates { (6.7,-8) (7.1,-8.1) (7.4,-6.9) (7.8,-7)};
    \draw [fill=cyan!30!white] (7.5,-8) rectangle (8.1,-7);
    \draw[->] (8.15,-7.5) -- (8.6,-7.5);
    \draw[->] (7.8,-8) -- (7.8,-8.25);
    \node at (7.8,-8.67) {\LARGE \textit{calorie}};
    
    \draw[thick,dotted,->] plot [smooth, tension=1.25] coordinates { (7.8,-8) (8.2,-8.1) (8.5,-6.9) (8.9,-7)};
    \draw [fill=cyan!30!white] (8.6,-8) rectangle (9.2,-7);
    \draw[->] (9.25,-7.5) -- (9.7,-7.5);
    \draw[->] (8.9,-8) -- (8.9,-8.25);
    \node at (9.05,-8.67) {\LARGE \textit{intake}};
    
    \draw[thick,dotted,->] plot [smooth, tension=1.25] coordinates { (8.9,-8) (9.3,-8.1) (9.6,-6.9) (10,-7)};
    \draw [fill=cyan!30!white] (9.7,-8) rectangle (10.3,-7);
    \draw[->] (10.35,-7.5) -- (10.8,-7.5);
    \draw[->] (10,-8) -- (10,-8.25);
    \node at (10,-8.67) {\LARGE \textit{has}};
    
    \draw[thick,dotted,->] plot [smooth, tension=1.25] coordinates { (10,-8) (10.4,-8.1) (10.7,-6.9) (11.1,-7)};
    \draw [fill=cyan!30!white] (10.8,-8) rectangle (11.4,-7);
    \draw[->] (11.45,-7.5) -- (11.9,-7.5);
    \draw[->] (11.1,-8) -- (11.1,-8.25);
    \node at (11.1,-8.67) {\LARGE \textit{been}};
    
    \draw[thick,dotted,->] plot [smooth, tension=1.25] coordinates { (11.1,-8) (11.5,-8.1) (11.8,-6.9) (12.2,-7)};
    \draw [fill=cyan!30!white] (11.9,-8) rectangle (12.5,-7);
    \draw[->] (12.55,-7.5) -- (13,-7.5);
    \draw[->] (12.2,-8) -- (12.2,-8.25);
    \node at (12.2,-8.67) {\LARGE \textit{low.}};
    
    \draw[thick,dotted,->] plot [smooth, tension=1.25] coordinates { (12.2,-8) (12.6,-8.1) (12.9,-6.9) (13.3,-7)};
    \draw [fill=cyan!30!white] (13,-8) rectangle (13.6,-7);
    \draw[->] (13.3,-8) -- (13.3,-8.25);
    \node at (13.3,-8.65) {\LARGE \textit{$<$/s$>$}};
    
    %% Decoder 2 %%
    \node at (1.2,-9.5) {\LARGE $<$s$>$};
    \draw[->] (1.2,-9.7) -- (1.2,-10);
    \draw [fill=purple!30!white] (0.9,-11) rectangle (1.5,-10);
    \draw[->] (1.55,-10.5) -- (2.15,-10.5);
    \draw[->] (1.2,-11) -- (1.2,-11.25);
    \node at (1.2,-11.65) {\LARGE \textit{In}};
    
    \draw[thick,dotted,->] plot [smooth, tension=1.25] coordinates { (1.2,-11) (1.6,-11.1) (1.9,-9.9) (2.3,-10)};
    \draw [fill=purple!30!white] (2,-11) rectangle (2.6,-10);
    \draw[->] (2.65,-10.5) -- (3.1,-10.5);
    \draw[->] (2.3,-11) -- (2.3,-11.25);
    \node at (2.3,-11.65) {\LARGE \textit{the}};
    
    \draw[thick,dotted,->] plot [smooth, tension=1.25] coordinates { (2.3,-11) (2.7,-11.1) (3,-9.9) (3.4,-10)};
    \draw [fill=purple!30!white] (3.1,-11) rectangle (3.7,-10);
    \draw[->] (3.75,-10.5) -- (4.2,-10.5);
    \draw[->] (3.4,-11) -- (3.4,-11.25);
    \node at (3.4,-11.71) {\LARGE \textit{past}};
    
    \draw[thick,dotted,->] plot [smooth, tension=1.25] coordinates { (3.4,-11) (3.8,-11.1) (4.1,-9.9) (4.5,-10)};
    \draw [fill=purple!30!white] (4.2,-11) rectangle (4.8,-10);
    \draw[->] (4.85,-10.5) -- (5.3,-10.5);
    \draw[->] (4.5,-11) -- (4.5,-11.25);
    \node at (4.5,-11.69) {\LARGE \textit{full}};
    
    \draw[thick,dotted,->] plot [smooth, tension=1.25] coordinates { (4.5,-11) (4.9,-11.1) (5.2,-9.9) (5.6,-10)};
    \draw [fill=purple!30!white] (5.3,-11) rectangle (5.9,-10);
    \draw[->] (5.95,-10.5) -- (6.4,-10.5);
    \draw[->] (5.6,-11) -- (5.6,-11.25);
    \node at (5.6,-11.65) {\LARGE \textit{TW}};
    
    \draw[thick,dotted,->] plot [smooth, tension=1.25] coordinates { (5.6,-11) (6,-11.1) (6.3,-9.9) (6.7,-10)};
    \draw [fill=purple!30!white] (6.4,-11) rectangle (7,-10);
    \draw[->] (7.05,-10.5) -- (7.5,-10.5);
    \draw[->] (6.7,-11) -- (6.7,-11.25);
    \node at (6.7,-11.71) {\LARGE \textit{your}};
    
    \draw[thick,dotted,->] plot [smooth, tension=1.25] coordinates { (6.7,-11) (7.1,-11.1) (7.4,-9.9) (7.8,-10)};
    \draw [fill=purple!30!white] (7.5,-11) rectangle (8.1,-10);
    \draw[->] (8.15,-10.5) -- (8.6,-10.5);
    \draw[->] (7.8,-11) -- (7.8,-11.25);
    \node at (7.8,-11.64) {\LARGE \textit{A}};
    
    \draw[thick,dotted,->] plot [smooth, tension=1.25] coordinates { (7.8,-11) (8.2,-11.1) (8.5,-9.9) (8.9,-10)};
    \draw [fill=purple!30!white] (8.6,-11) rectangle (9.2,-10);
    \draw[->] (9.25,-10.5) -- (9.7,-10.5);
    \draw[->] (8.9,-11) -- (8.9,-11.25);
    \node at (8.9,-11.64) {\LARGE \textit{A}};
    
    \draw[thick,dotted,->] plot [smooth, tension=1.25] coordinates { (8.9,-11) (9.3,-11.1) (9.6,-9.9) (10,-10)};
    \draw [fill=purple!30!white] (9.7,-11) rectangle (10.3,-10);
    \draw[->] (10.35,-10.5) -- (10.8,-10.5);
    \draw[->] (10,-11) -- (10,-11.25);
    \node at (10,-11.64) {\LARGE \textit{has}};
    
    \draw[thick,dotted,->] plot [smooth, tension=1.25] coordinates { (10,-11) (10.4,-11.1) (10.7,-9.9) (11.1,-10)};
    \draw [fill=purple!30!white] (10.8,-11) rectangle (11.4,-10);
    \draw[->] (11.45,-10.5) -- (11.9,-10.5);
    \draw[->] (11.1,-11) -- (11.1,-11.25);
    \node at (11.1,-11.64) {\LARGE \textit{been}};
    
    \draw[thick,dotted,->] plot [smooth, tension=1.25] coordinates { (11.1,-11) (11.5,-11.1) (11.8,-9.9) (12.2,-10)};
    \draw [fill=purple!30!white] (11.9,-11) rectangle (12.5,-10);
    \draw[->] (12.55,-10.5) -- (13,-10.5);
    \draw[->] (12.2,-11) -- (12.2,-11.25);
    \node at (12.2,-11.65) {\LARGE \textit{S}};
    
    \draw[thick,dotted,->] plot [smooth, tension=1.25] coordinates { (12.2,-11) (12.6,-11.1) (12.9,-9.9) (13.3,-10)};
    \draw [fill=purple!30!white] (13,-11) rectangle (13.6,-10);
    \draw[->] (13.3,-11) -- (13.3,-11.25);
    \node at (13.3,-11.65) {\LARGE \textit{$<$/s$>$}};
    \end{tikzpicture}
    }
    \vspace{-0.15in}
    \caption{\small Decoder outputs: (blue) natural language summary, and (pink) protoform template.}
    \label{fig:output}
    \vspace{-0.2in}
\end{figure}
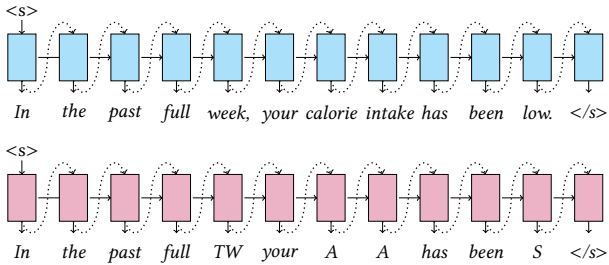